%% file: example_paper.tex
\documentclass{article}

\usepackage[main,final]{neurips_2026}
\usepackage{microtype}
\usepackage{graphicx}
\usepackage{subcaption}
\usepackage{booktabs} 
\usepackage{hyperref}
\usepackage{inconsolata}
\usepackage{amsmath}
\usepackage{amssymb}
\usepackage{mathtools}
\usepackage{amsthm}
\usepackage{xspace}
\usepackage{algorithm}
\usepackage{algpseudocode}

\usepackage[capitalize,noabbrev]{cleveref}
\usepackage{listings}
\usepackage{xcolor}
\usepackage[textsize=tiny]{todonotes}
\usepackage{natbib}
\usepackage{multirow}

\definecolor{codegreen}{rgb}{0,0.6,0}
\definecolor{codegray}{rgb}{0.5,0.5,0.5}
\definecolor{codepurple}{rgb}{0.58,0,0.82}
\definecolor{backcolour}{rgb}{0.95,0.95,0.92}

\lstdefinestyle{mystyle}
{
    backgroundcolor=\color{backcolour},    
    commentstyle=\color{codegreen},
    keywordstyle=\color{magenta},
    numberstyle=\tiny\color{codegray},
    stringstyle=\color{codepurple},
    basicstyle=\ttfamily\footnotesize, 
    breakatwhitespace=false,          
    breaklines=true,                  
    captionpos=t,                     
    keepspaces=true,                  
    numbers=left,                     
    numbersep=5pt,                   
    showspaces=false,                 
    showstringspaces=false,
    showtabs=false,                   
    tabsize=4,
    upquote=true,
    literate={"}{{\fontencoding{T1}\selectfont\char34}}1
             {~}{{\char126}}1
}
\theoremstyle{plain}

\theoremstyle{definition}

\theoremstyle{remark}

\newcommand{\sys}{Qrita\xspace}
\newcommand{\topk}{Top-$k$\xspace}
\newcommand{\topp}{Top-$p$\xspace}

\title{\sys: High-performance \topk and \topp Algorithm for GPUs
using Pivot-based Truncation and Selection}

\author{
  Jongseok Park\\
  University of California, Berkeley\\
  \texttt{js\_park@berkeley.edu} \\
  \And
  Sunga Kim\\
  University of California, Berkeley\\
  \texttt{sunga.kim@berkeley.edu} \\
  \And
  Alvin Cheung\\
  University of California, Berkeley\\
  \texttt{akcheung@berkeley.edu} \\
  \And
  Ion Stoica\\
  University of California, Berkeley\\
  \texttt{istoica@berkeley.edu} \\
}

\begin{document}

\maketitle

\begin{abstract}
Despite their importance in model sampling, efficient implementation of the \topk and \topp algorithms poses a significant challenge as model vocabulary grows with each new model release. Existing approaches often rely on sorting, which incurs significant computation and memory overhead on GPUs, or on stochastic approaches that alter the algorithm's output. In this work, we propose \sys, an efficient \topk and \topp algorithm based on pivot-based truncation and selection. \sys leverages pivot-based search for both \topk and \topp with two key techniques: 1. Gaussian-based $\sigma$-truncation, which \textbf{limits the complexity of the algorithm to $\mathcal{O}(n)$}, and 2. Quaternary pivot search with duplication handling, which halves the number of pivot search iterations and \textbf{guarantees deterministic output}. We implement \sys using Triton and evaluate its performance against the \topk and \topp kernels of high-performance LLM execution engines such as SGLang and FlashInfer, improving end-to-end serving throughput up to 1.4$\times$ with half the memory usage, while providing the \textbf{same output as stable sort-based \topk and \topp algorithms}. \sys is now the default Triton \topk and \topp sampler of vLLM, and its implementation is available at \url{https://github.com/vllm-project/vllm/blob/main/vllm/v1/sample/ops/topk_topp_triton.py}.
\end{abstract}

\newif\ifcomments
\commentstrue
\ifcomments
    \providecommand{\alvin}[1]{{\protect\color{black}{\bf [alvin: #1]}}}
    \providecommand{\jongseok}[1]{{\protect\color{black}{\bf [Jongseok: #1]}}}
    \providecommand{\sunga}[1]{{\protect\color{black}{\bf [Sunga: #1]}}}
    \providecommand{\ion}[1]{{\protect\color{black}{\bf [Ion: #1]}}}
\else
    \providecommand{\alvin}[1]{}
    \providecommand{\jongseok}[1]{}
    \providecommand{\sunga}[1]{}
    \providecommand{\ion}[1]{}
\fi

\input{1.Introduction}
\input{2.Related_Work}
\input{3.Method}

\input{4.Evaluations}

\input{5.Discussions}

\input{6.Conclusion}

\bibliography{example_paper}
\bibliographystyle{plain}


\end{document}

%% file: 1.Introduction.tex
\section{Introduction}
\label{sec:intro}

\topk and \topp truncation are two key mechanisms used in sampling-based generation of Large Language Models (LLMs) to balance output quality and diversity.
\topk~\cite{fan2018hierarchicalneuralstorygeneration} selects the $k$ highest-logit tokens, pruning low-probability options to prevent incoherent output. In contrast, \topp~\citep{holtzman2020curiouscaseneuraltext} selects the smallest subset whose cumulative probability exceeds a threshold $p$, offering a dynamic and diversity-oriented cutoff. Together, \topk and \topp provide a robust solution that stabilizes LLM generation while preserving contextual variation.

Despite their conceptual simplicity, implementing efficient \topk and \topp algorithms on GPUs remains a significant challenge. Modern LLM execution engines, such as vLLM~\citep{kwon2023efficientmemorymanagementlarge}, rely on full-vocabulary sort-and-filter methods to identify the target \topk and \topp subset. Such sorting-based algorithms provide exact results but are computationally inefficient on GPUs. Divide-and-conquer solutions such as the Bitonic sort~\cite{satish2009designing} provide ways to utilize the massively parallel architecture of GPUs, but still incur significant memory overhead as repeated comparisons between array elements require strided memory accesses and substantial buffer space for intermediate results.

Recent research explores alternative directions to mitigate these inefficiencies. FlashInfer~\citep{ye2025flashinferefficientcustomizableattention} eliminates sorting overhead via iterative \textit{rejection sampling}, designed to be stochastically equivalent to \topk and \topp sampling. However, because it returns a single token with only probabilistic guarantees, it is incompatible with tasks that require the full deterministic candidate set, such as speculative decoding verification~\cite{leviathan2023fast} and reinforcement learning~\cite{nagarajan2018impact, he2025nondeterminism}. Other solutions, such as Top-$n\sigma$~\citep{tang2024topnsigmalogitsneed} and Min-$p$~\cite{nguyen2025turningheatminpsampling}, propose statistical cutoffs based purely on the logit distribution. Statistical cutoffs remove sorting entirely, allowing for highly efficient single-pass $\mathcal{O}(n)$ algorithms. However, the generation quality of heuristic methods is highly debated~\cite{schaeffer2025minpmaxexaggerationcritical}, and standard \topk and \topp sampling remain the dominant method in production environments~\cite{kwon2023efficientmemorymanagementlarge, zheng2024sglang}.

Thus, a critical gap remains for a \topk and \topp algorithm that delivers both memory efficiency and strict deterministic behavior on the ever-increasing LLM vocabulary. To this end, we investigate R\topk~\citep{xie2025rtopkultrafastrowwisetopk}, a pivot-based truncation strategy originally designed for node selection in Graph Neural Networks. R\topk offers a memory-efficient alternative to sorting using an in-place binary search for a \topk \textbf{truncation pivot}, a value where exactly $k$ elements in the set are larger than or equal to it. However, R\topk is not directly applicable to LLM sampling for two reasons: (1) its search algorithm is optimized for small vectors ($<$1,000 elements) and becomes inefficient over the large LLM vocabularies often exceeding 100,000 elements; and (2) it lacks explicit handling for duplicate logits, which can trigger excessive search iterations and non-deterministic behavior.

To address these limitations, we propose \sys, a high-performance \topk and \topp algorithm for GPUs. \sys extends the concept of pivot-based truncation of R\topk to be compatible with the joint \topk and \topp pipeline in LLM decoding, employing two key techniques to achieve computational efficiency and determinism: (1) \textbf{Gaussian $\sigma$-truncation}, limits the which \textit{complexity of the algorithm to $\mathcal{O}(n)$} by leveraging the stochastic distribution of the logit values and the high sparsity of \topk and \topp operations in LLM sampling, and (2) \textbf{Quaternary pivot search with duplication handling}, which halves the search iterations while guaranteeing determinism with \textit{exact same output as stable sort-based \topk and \topp algorithms} using multi-pivot bookkeeping of duplicate logit values, allowing fast and reproducible outputs.

In summary, \sys provides fast and deterministic Top-$k$/Top-$p$ via truncation, pivot search, and duplication handling, with \textbf{three key advantages} over existing algorithms. (1) \sys avoids full-vocabulary sorting, reducing memory overhead and providing high performance on GPUs. (2) \sys returns the full Top-$k$/Top-$p$ candidate set instead of a single sample token, retaining compatibility with downstream tasks that require the full set. (3) \sys retains exact Top-$k$/Top-$p$ semantics and reproducibility, unlike statistical cutoff methods that alter the sampling output.

We implement \sys using Triton~\cite{tillet2019triton}, a popular GPU programming language. We provide ablation studies of \sys and perform extensive evaluations against the state-of-the-art solutions provided by vLLM~\cite{kwon2023efficientmemorymanagementlarge}, SGLang~\cite{zheng2024sglang}, and FlashInfer~\cite{ye2025flashinferefficientcustomizableattention}, demonstrating throughput improvements up to 1.4$\times$ while providing the same output as the stable sorting-based \topk and \topp algorithms.



    

%% file: 2.Related_Work.tex
\section{Related Work}
\vspace{-1pt}
\label{sec:related_work}

\noindent{\textbf{\topk and \topp Sampling in LLMs:}}
\topk~\cite{fan2018hierarchicalneuralstorygeneration} and \topp (nucleus) sampling~\cite{holtzman2020curiouscaseneuraltext} are the standard mechanisms for controlling the randomness of LLM generation. \topk restricts the candidate pool to the $k$ most probable tokens, effectively filtering out the long tail of low-probability tokens. \topp refines this approach by dynamically truncating the tail based on the cumulative probability mass $p$, allowing the candidate set size to expand or contract based on the model's confidence. These operators are typically implemented by sorting the entire vocabulary (often $>100,000$ tokens), truncating the top candidates, and/or calculating the cumulative sum of probabilities~\cite{kwon2023efficientmemorymanagementlarge}. While true to the definition of \topk and \topp, this sorting-based approach is inefficient on the massively parallel architecture of GPUs and creates a significant memory overhead in the form of strided memory access and intermediate output buffers.

High-performance libraries such as FlashInfer~\citep{ye2025flashinferefficientcustomizableattention} provide sorting-free \topk and \topp sampling by leveraging Dual Pivot Rejection Sampling, an algorithm that iteratively refines the inverse transform probability range with rejection sampling until a token is sampled~\cite{xing2025sortingfree}.
While this design returns a stochastically equivalent result to the original \topk and \topp-based sampling, it fundamentally alters the operator's interface: it returns a \textit{single sampled index} rather than a set of candidates. This limitation renders it inadequate for advanced decoding tasks that require inspecting the candidate set, such as speculative decoding verification~\cite{leviathan2023fast} or semantic caching systems~\cite{bang2023gptcache}, where a known ``best'' set is required. Furthermore, as the parallel prefix-sum operations required for inverse transform sampling are non-associative in floating-point arithmetic, non-deterministic behavior and numerical instability may arise when executed across different hardware configurations~\cite{xing2025sortingfree}.

\noindent{\textbf{Statistical Threshold Methods:}}
An alternative direction, explored by methods such as $\eta$-sampling~\cite{hewitt2022truncationsamplinglanguagemodel}, Top-$n\sigma$~\cite{tang2024topnsigmalogitsneed}, and Min-$p$~\cite{nguyen2025turningheatminpsampling}, is to eliminate sorting by relying on the statistical properties of the logit distribution. These approaches observe that LLM logits typically follow a quasi-Gaussian distribution, where a large number of ``noisy'' tokens cluster around the mean and a small number of ``informative'' tokens form a high-probability outlier group. By calculating simple statistics like the mean ($\mu$) and standard deviation ($\sigma$) of the distribution, these methods define a hard cutoff threshold (e.g., $\tau = \mu + n\sigma$) to separate the signal from the noise in a single pass.
While computationally efficient, these statistical methods produce a completely different sampling output compared to standard \topk and \topp, leading to debated effects on generation coherence~\cite{schaeffer2025minpmaxexaggerationcritical}. Our work leverages this statistical intuition not as a replacement, but as a \textit{pre-filtering} step (Gaussian $\sigma$-truncation) to \topk and \topp to accelerate the algorithm without altering the final output.

\noindent{\textbf{Pivot-Based Selection Algorithm:}}
Quickselect~\cite{hoare1961algorithmquickselct} and more recent algorithms such as RadixSelect~\cite{alabi2012fastradixquick} and R\topk~\citep{xie2025rtopkultrafastrowwisetopk} replace sorting with a search and select of a $k$-th pivot value. By iteratively selecting a pivot value $\tau$ and counting the number of elements $x_i \ge \tau$, the algorithm can narrow down the range containing the $k$-th largest element without moving data.
This approach is highly effective for small vector sizes where the data fits in fast on-chip memory. However, for LLMs with massive vocabularies, the repeated passes across the logit set on each iteration saturate the memory bandwidth, making it slower than optimized solutions such as that of FlashInfer. Furthermore, standard pivot-based methods are non-deterministic in the presence of duplicate logits often found in long-tailed distributions with low precision, as the search cannot detect and distinguish indices with duplicate values. \sys addresses these limitations by using a quaternary search to reduce memory passes and bookkeeping of duplicate logits to ensure determinism.

%% file: 3.Method.tex
\section{Algorithm Design}
\label{sec:method}


    
    
    

\begin{algorithm}[t]
\caption{High-level \sys overview for \topk}
\small
\label{alg:high_level}
\begin{algorithmic}[1]
\State \textbf{Input:} Batched logits $\mathbf{Z}$, \topk target $\mathbf{K}$, Vocab size $\mathcal{V}$, Batch size $B$
\State \textbf{Output:} \topk output $\mathbf{O}$
\For{$b$ in $0 \ldots B-1$} \Comment{Parallelized as Triton Programs}
    \State $\mathbf{Z}_{t} \leftarrow \textsc{GaussianSigmaTruncation}(\mathbf{Z}[b], \mathcal{V})$ \Comment{Truncate Gaussian noise region}
    \If{$|\mathbf{Z}_{t}| > \mathbf{K}[b]$} \Comment{Truncation Hit: Truncated set large enough}
        \State $\mathbf{Z}_{s} \leftarrow \mathbf{Z}_{t}$ \Comment{Search truncated set}
    \Else \Comment{Truncation Miss: Not enough elements}
        \State $\mathbf{Z}_{s} \leftarrow \mathbf{Z}[b]$ \Comment{Search original full logit set}
    \EndIf
    \State $\tau, z_{dup}, N_{dup} \leftarrow \textsc{QuaternarySearch}(\mathbf{Z}_{s}, \mathbf{K}[b])$ \Comment{Search pivot and find duplicates}
    \State $\mathbf{O}[b] \leftarrow \textsc{DuplicationHandling}(\mathbf{Z}, \tau, z_{dup}, N_{dup})$ \Comment{Resolve duplicates for exact Top-k}
\EndFor
\end{algorithmic}
\end{algorithm}

We present the design of \sys using \topk as the main example, and highlight how the \topp algorithm differs from \topk. Algorithm \ref{alg:high_level} provides a high-level overview of \sys. The {\tt for} loop across the batch dimension (line 3) is parallelized as Triton Programs\footnote{A Triton Program is equivalent to a CUDA Threadblock or a ROCm Workgroup.}~\cite{triton_language_api}, and operations inside each batch are parallelized as Triton blocks. We use a persistent kernel design and launch $\min(B, N_{sm})$ Triton programs, where $N_{sm}$ is the number of multiprocessors on the GPU. In the following, we skip the batch dimension subscripts when explaining the operations within each batch for brevity. The equality checks $a = b$ in algorithms are performed using $abs(a-b) < 10^{-9}$ in implementation.

\subsection{Gaussian \texorpdfstring{$\sigma$}{sigma}-truncation}

Although the vocabulary size of modern LLMs is on the order of 100k entries, final predictions are typically concentrated within fewer than 100 vocabulary entries per token, regardless of the model used or modifications such as finetuning or quantization. These few high-probability entries form the outlier region of the logit distribution, while the vast majority of the vocabulary forms Gaussian noise~\cite{tang2024topnsigmalogitsneed}. The goal of Gaussian-based $\sigma$-truncation is to truncate this massive noise region in a \textbf{single linear pass}, while its large headroom of a few hundred vocabulary entries allows reliable capturing and forwarding of the real predictions to the quaternary search for exact \topk and \topp sampling. This step limits the \textbf{complexity of \sys{} to $\mathcal{O}(n)$}, providing a much more efficient bound over the ever-growing LLM vocabulary compared to existing $\mathcal{O}(n\log(n))$ \topk and \topp algorithms.

Let $\mathbf{Z} \in \mathbb{R}^{\mathcal{V}}$ denote the model logits. We first approximate the distribution of logits as a quasi-Gaussian $\mathbf{Z} \sim \mathcal{N}(\mu, \sigma^2)$. We sample a block of logits and compute the sample mean $\mu$ and standard deviation $\sigma$ of the logits to determine a truncation threshold $t = \mu + \delta \cdot \sigma$.
We use a lookup table to quickly approximate the coefficient $\delta$ from the target $k$ or $p$ value. We precompute table entries using the normal quantile function $\Phi_k^{-1}$ where $\delta_k = \Phi_k^{-1}(1 - k/\mathcal{V})$ for $k$ and the inverse cumulative softmax probability of the samples $\Phi_p^{-1}$ where $\delta_p = \Phi_p^{-1}(p)$.

\begin{algorithm}[t]
\caption{Quaternary \topk Pivot Search}
\small
\label{alg:quaternary_top_k}
\begin{algorithmic}[1]
\State \textbf{Input:} Search set $\mathbf{Z_s}$, \topk target $k$
\State \textbf{Output:} Pivot $\tau,\;z_{dup},\;N_{dup}$
\State $\tau \leftarrow \infty,\;z_{dup} \leftarrow 0,\;N_{dup} \leftarrow 0$ \Comment{Initialize output values}
\State $l \leftarrow \min(Z_s),\;r \leftarrow \max(Z_s),\;i \leftarrow 0$ \Comment{Initialize search range}
\While{$(|r-l| \geq 10^{-9})\ \wedge\ (i < 18)$} \Comment{Iterate until range converges or max iterations}
    \For{$n$ in $1, 2, 3$} \Comment{Evaluate three pivots (quaternary search)}
        \State $\tau_n \leftarrow l + (n/4)\times(r-l)$ \Comment{Set pivots at 25th, 50th, 75th percentile}
        \State $N_{\tau_n} \leftarrow |\{ z_i \;\mid\; z_i > \tau_n \}|$ \Comment{Count elements above pivot}
        \State $z_{mn} \leftarrow \min (\{ z_i \mid z_i > \tau_n \})$ \Comment{Track minimum logit above pivot}
        \State $N_{z_{mn}} \leftarrow |\{ z_i \mid z_i = z_{mn} \}|$ \Comment{Count occurrences of minimum logit}
        \If{$(N_{\tau_n} \geq k) \wedge\ (N_{\tau_n} - N_{z_{mn}} < k)$} \Comment{Termination: Pivot straddles $k$ with duplicates}
            \State $\tau \leftarrow \tau_n,\;z_{dup} \leftarrow z_{mn},\;N_{dup} \leftarrow N_{z_{mn}}$ \Comment{Record pivot and duplicate info}
            \State \textbf{break}
        \EndIf
    \EndFor
    \State $r \leftarrow (N_{\tau_1} < k)?\;\tau_1 : (N_{\tau_2} < k)?\;\tau_2 : (N_{\tau_3} < k)?\;\tau_3:r$ \Comment{Narrow upper bound}
    \State $l \leftarrow (N_{\tau_3} > k)?\;\tau_3 : (N_{\tau_2} > k)?\;\tau_2 : (N_{\tau_1} > k)?\;\tau_1:l$ \Comment{Narrow lower bound}
    \State $i \leftarrow i + 1$        \Comment{Increment iteration}
    \State $\tau \leftarrow (l + r)/2$ \Comment{Set fallback pivot to midpoint}
\EndWhile
\State \textbf{return} $(\tau,\;z_{dup},\;N_{dup})$
\end{algorithmic}
\end{algorithm}

\begin{figure}[t]
\vspace{-8pt}
    \centering
    \begin{subfigure}[b]{0.27\linewidth}
        \includegraphics[width=\linewidth]{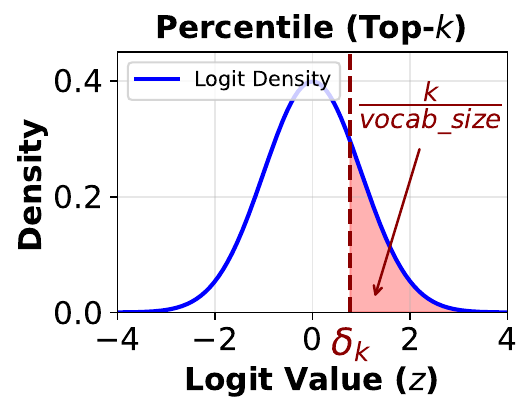}
    \end{subfigure}
    \begin{subfigure}[b]{0.27\linewidth}
        \includegraphics[width=\linewidth]{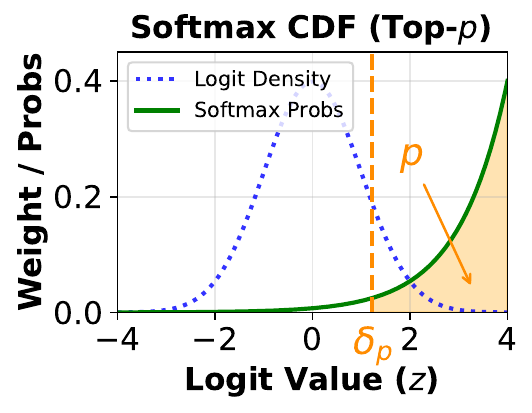}
    \end{subfigure}
    \caption{Gaussian $\sigma$-truncation on logits following a standard normal distribution.}
    \label{fig:gaussian_trunc}
    \vspace{-16pt}
\end{figure}

Figure~\ref{fig:gaussian_trunc} illustrates Gaussian $\sigma$-truncation on logits following a standard normal distribution ($t = \delta$). In practice, 
we subtract $0.2\times abs(t)$ from $t$ as a safety margin and gather the truncated \textit{outlier set} that satisfies $\mathbf{Z}_t = \left\{ z_i \;\middle|\; z_i > t \right\}$. We allocate additional memory for $\mathcal{V}$ elements per Triton program to store the truncated set. If $|\mathbf{Z}_{t}| > k$, or a ``truncation hit,'' we pass the pointer for $\mathbf{Z}_t$ to the Quaternary Search; if not, we fall back and pass the pointer for $\mathbf{Z}$.
For \topp, we compute the softmax of the logits in this stage.\footnote{Softmax on the unsorted logits may produce different output to softmax after sorting due to floating point non-associativity.} We gather and sum the \textit{outlier probabilities}, and check if the \textit{sum} is larger than $p$ instead. We provide details on the benefits of Gaussian $\sigma$-truncation and its hit-rate in Section~\ref{sec:mem_and_hitrate}.

\subsection{Quaternary Pivot Search}
Even with the Gaussian $\sigma$-truncation, the overhead of pivot search is still significant. To mitigate this, we use the Quaternary Pivot Search in Algorithm~\ref{alg:quaternary_top_k}. We increase the number of pivots to three to halve the number of iterations and track the occurrence of the smallest logit that is larger than the pivot. This prevents the search from falling into an excessive number of iterations and nondeterministic behavior in the presence of duplicate logits, to be explained later.

We first set our initial search range $r \leftarrow \max(\mathbf{Z}_s)$ and $l \leftarrow\min(\mathbf{Z}_s)$ (line 4), where $\max(\mathbf{Z}_s)$ is $\max(\mathbf{Z})$ and $\min(\mathbf{Z}_s)$ is $t$ if the truncation hits and $\min(\mathbf{Z})$ if not. For each iteration, we set the search pivots $\tau_n$ ($n=1,2,3$) to the 25th, 50th, 75th percentiles of the search range (line 7) and count $N_{\tau_n} = |\left\{ z_i \;\mid\; z_i > \tau_n \right\}|$ (line 8). We also track the \textit{minimum logit} $z_{mn}$ that is larger than $\tau_n$, or $z_{mn} = \min(\{z_i\mid\ z_i > \tau_n\})$ (line 9), and its occurrence $N_{z_{mn}} = |\left\{ z_i \;\middle|\; z_i = z_{mn} \right\}|$ (line 10).

The search ends when $(N_{\tau_n} \geq k) \wedge\ (N_{\tau_n} - N_{z_{mn}} < k)$ (line 11), which finds the pivot $\tau_n$ where the exclusion of $Z_{mn}$ logits pushes $N_{\tau_n}$ below $k$. This provides a robust termination condition for the pivot search, as it detects the case where multiple $Z_{mn}$ logits exist and must be partially discarded to satisfy \topk. Without this, the naive $N_{\tau_n} = k$ condition will fall into an infinite loop. Once this condition is met, we return $\tau_n$, $z_{mn}$, and $N_{z_{mn}}$, now labeled as $\tau$, $z_{dup}$, and $N_{dup}$. If the condition is not met for any of the pivots, we update the search range. If $N_{\tau_1} < k$, this means that the true $\tau < \tau_1$, therefore $r \leftarrow \tau_1$. Similarly, if $N_{\tau_3} > k$, $\tau > \tau_3$, therefore $l \leftarrow \tau_3$. We chain the update as in lines 16 and 17 of Algorithm~\ref{alg:quaternary_top_k}, and iterate until the search range $|r - l|$ falls below the tolerance of $10^{-9}$, or up to 18 iterations, which reduces the range to $1/4^{18} = 1/2^{36}$ of the original. 

 For \topp, we use the same algorithm but with pivot probability $\pi_n$, and calculate the sum of probabilities $S_{\pi_n} = \sum_{p_i > \pi_n} p_i$ while tracking $p_{mn} = \min(\{p_i \mid\ p_i > \pi_n\})$ and $S_{p_{mn}} = \sum_{p_i = p_{mn}} p_i$. The search terminates when $(S_{\pi_n} \geq p) \wedge\ (S_{\pi_n} - S_{p_{mn}} < p)$, and we still count $N_{p_{mn}}$ to be later used for duplication handling. For range updates, we do $r \leftarrow (S_{\pi_1} < p)?\pi_1 : (S_{\pi_2} < p)?\pi_2 : (S_{\pi_3} < p)?\pi_3:r$ and $l \leftarrow (S_{\pi_3} > p)?\pi_3 : (S_{\pi_2} > p)?\pi_2 : (S_{\pi_1} > p)?\pi_1:l$.
 
 We also experiment with a larger number of pivots. However, due to the increased pressure on hardware resources such as the register file, the per-iteration time increases significantly to a point where it outweighs the benefits from reduced iterations. We provide details on the tradeoff between binary and quaternary search in Section~\ref{sec:ablation}.

\subsection{Duplication Handling}

\begin{algorithm}[t]
\small
\caption{\topk output with Duplication Handling}
\label{alg:dup}
\begin{algorithmic}[1]
\State \textbf{Input:} Logits set $\mathbf{Z}$, $\tau,\;z_{dup},\;N_{dup}$
\State \textbf{Output:} Output $\mathbf{O}$
\State $\mathbf{M}_\tau \leftarrow 1(\mathbf{Z} > \tau)$ \Comment{Base mask: all logits above pivot}
\State $\mathbf{N}_\tau \leftarrow \Sigma M_{\tau}$    \Comment{Number of all logits above pivot}
\State $N_{keep} \leftarrow N_{dup} - (N_{\tau} - k)$ \Comment{Number of duplicate logits to retain}
\State $\mathbf{M}_{z_{dup}} \leftarrow 1(\mathbf{Z} = z_{dup})$ \Comment{Locate all occurrences of the duplicate boundary logit}
\State $\mathbf{C}_{z_{dup}} \leftarrow cumsum(\mathbf{M}_{z_{dup}}) \wedge\ M_{z_{dup}}$ \Comment{Occurrence map: Index duplicates in appearance order}
\State $\mathbf{M}_{remv} \leftarrow 1(\mathbf{C}_{z_{dup}} > N_{keep})$ \Comment{Mark excess duplicates for removal}
\State $\mathbf{M}_{final} \leftarrow \mathbf{M}_\tau \wedge\ \neg\mathbf{M}_{remv}$ \Comment{Final mask: Base set minus removed duplicates}
\State $\mathbf{O} \leftarrow \mathbf{Z} + (1 - \mathbf{M}_{final}) \cdot (-\infty)$ \Comment{Apply mask in-place; masked logits set to $-\infty$}
\State \textbf{return} $\mathbf{O}$
\end{algorithmic}
\end{algorithm}

Algorithm~\ref{alg:dup} shows the final output generation with duplication handling. First, we create a mask $\mathbf{M}_\tau = 1(\mathbf{Z} > \tau)$ (line 3), as our base set output logits. As $\mathbf{M}_\tau$ contains all duplicate logits, we must partially zero out the duplicate logits. We calculate the number of duplicate logits that we keep using $N_{keep} = N_{dup} - (N_{\tau} - k)$ for \topk, and $N_{keep} = N_{dup} - \lfloor (S_{\tau} - p) / p_{mn}\rfloor$ for \topp (line 5). To efficiently locate and remove unwanted duplicate logits in parallel, we first create a mask of duplicate logits $\mathbf{M}_{z_{dup}}$ (line 6) and use cumulative sum on the mask and perform logical AND with the original mask to create the \textit{occurrence map} of the duplicate logits, $\mathbf{C}_{z_{dup}}$ (line 7). We then create a mask of duplicate logits we remove, $\mathbf{M}_{remv}$, by comparing the occurrence map $\mathbf{C}_{z_{dup}}$ with $N_{keep}$ (line 8). Then we perform logical AND between $\mathbf{M}_\tau$ and $\neg\mathbf{M}_{remv}$ to get our final mask, $\mathbf{M}_{final}$ (line 9).

$\mathbf{M}_{final}$ maps whether each logit of $Z$ is included in our final \topk or \topp output. It can be used to gather the final output or mask out unwanted logits. As we support execution with different $k$ or $p$ values per batch dimension, we mask out unwanted logits in-place from $Z$ (line 10), and return our final output $O$. As we calculate the exact number of duplicate logits that must be kept, and always keep the ones that appear earlier in the logits set, our algorithm keeps the correct semantics of stable sort-based \topk and \topp algorithms with reproducible and deterministic outputs across all runs. 

\subsection{Tiered Triton Block Size Autotune}

We also include a Triton-specific optimization in our implementation. The vectorized operators of our algorithm, such as $\min()$, $\max()$, masking, or $cumsum()$, are executed using \textit{Triton blocks}. Triton splits the vector into fixed-size blocks, and the block elements are executed in parallel across GPU threads. 
If the blocks are too small, multiple iterations are needed to cover the whole vector, and if the blocks are too large, the blocks are padded, reducing hardware utilization. 
However, in \sys, the size of the vector operands differs greatly depending on whether the Gaussian $\sigma$-truncation is a hit or not. Therefore, we create tiered block sizes, where the larger block size is used for operations on the full logit set, and the smaller block size is used for operations on the truncated set. We use Triton Autotune to let the compiler find the best combination of block sizes. We show the benefit of our multi-block autotune in Section~\ref{sec:ablation}.

%% file: 4.Evaluations.tex
\section{Evaluation}
\label{sec:eval}

\subsection{Evaluation Setup}
\label{sec:eval_setup}
We evaluate \sys by comparing its performance against four popular baselines: a naive PyTorch~\cite{paszke2019pytorch} sort-based \topk and \topp implementation, vLLM~\cite{kwon2023efficientmemorymanagementlarge}\footnote{We compare to vLLM's default sampler before \sys.}, FlashInfer~\cite{ye2025flashinferefficientcustomizableattention}, and SGLang~\cite{zheng2024sglang}. Note that FlashInfer and SGLang return a single output token, which is stochastically equivalent to sampling from \topk or \topp. As such, they have a much more relaxed output constraint and are not equivalent to the strict definition of \topk or \topp. However, we still include them as baselines as they are the current state-of-the-art solutions for \topk and \topp sampling. 
We use samples from the WikiText-2~\cite{merity2016pointer} dataset to generate the input logits using five different LLMs, R1-Distill-Llama-8B~\cite{grattafiori2024llama3herdmodels, deepseekai2025deepseekr1incentivizingreasoningcapability}, Qwen3-8B~\cite{yang2025qwen3technicalreport}, GPT-OSS-20B~\cite{openai2025gptoss120bgptoss20bmodel}, Qwen3.5-35B-A3B~\cite{qwen3.5}, and Gemma3-4B~\cite{gemmateam2025gemma3technicalreport}, which have vocabulary sizes of 128256, 151936, 201088, 248320, and 262208 respectively. 
All experiments are executed on an NVIDIA H100 GPU, unless stated otherwise. 

\subsection{Performance}
\label{sec:performance}

\begin{figure*}[t]
  \centering

  \hspace*{0.02\textwidth}
  \begin{minipage}{0.24\textwidth}
    \centering\textbf{$k$=10, $p$=None}
  \end{minipage}\hfill
  \begin{minipage}{0.24\textwidth}
    \centering\textbf{$k$=None, $p$=0.7}
  \end{minipage}\hfill
  \begin{minipage}{0.24\textwidth}
    \centering\textbf{$k$=50, $p$=0.9}
  \end{minipage}\hfill
  \begin{minipage}{0.24\textwidth}
    \centering\textbf{$k$=Rand, $p$=Rand}
  \end{minipage}


  \begin{minipage}[c]{0.02\textwidth}
    \centering
    \rotatebox{90}{\small\textbf{Vocab=128256}}
  \end{minipage}%
  \hfill
  \begin{subfigure}[c]{0.24\textwidth}
    \includegraphics[width=\linewidth]
    {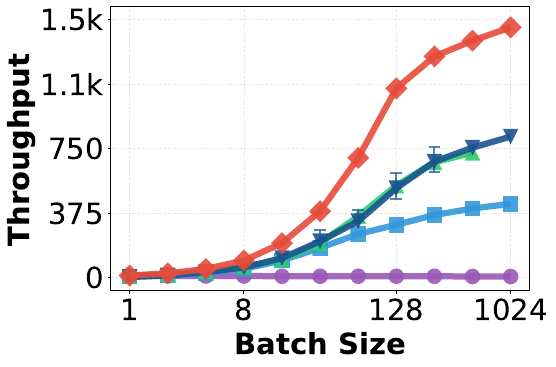}
  \end{subfigure}\hfill
  \begin{subfigure}[c]{0.24\textwidth}
    \includegraphics[width=\linewidth]
    {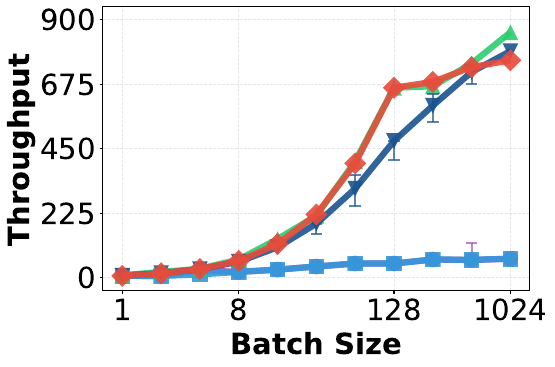}
  \end{subfigure}\hfill
  \begin{subfigure}[c]{0.24\textwidth}
    \includegraphics[width=\linewidth]
    {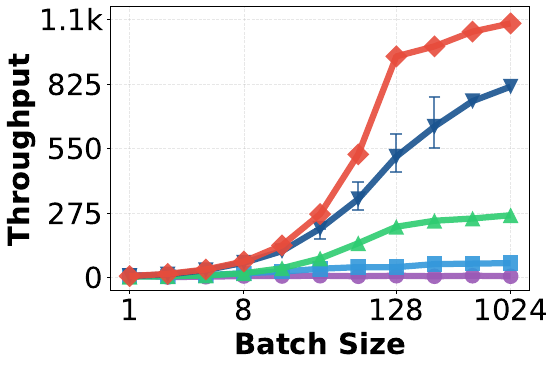}
  \end{subfigure}\hfill
  \begin{subfigure}[c]{0.24\textwidth}
    \includegraphics[width=\linewidth]
    {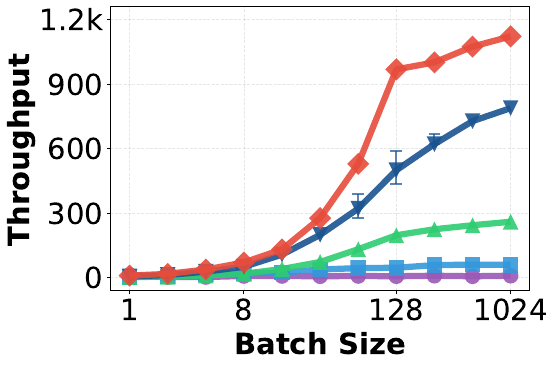}
  \end{subfigure}

  \begin{minipage}[c]{0.03\textwidth}
    \centering
    \rotatebox{90}{\small\textbf{Vocab=151936}}
  \end{minipage}%
  \hfill
  \begin{subfigure}[c]{0.24\textwidth}
    \includegraphics[width=\linewidth]
    {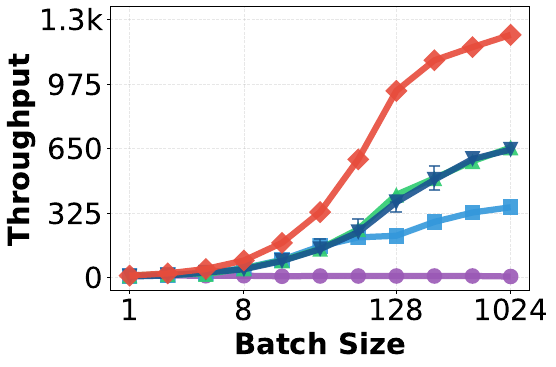}
  \end{subfigure}\hfill
  \begin{subfigure}[c]{0.24\textwidth}
    \includegraphics[width=\linewidth]
    {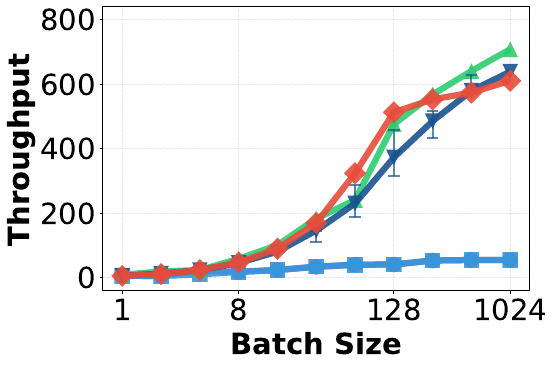}
  \end{subfigure}\hfill
  \begin{subfigure}[c]{0.24\textwidth}
    \includegraphics[width=\linewidth]
    {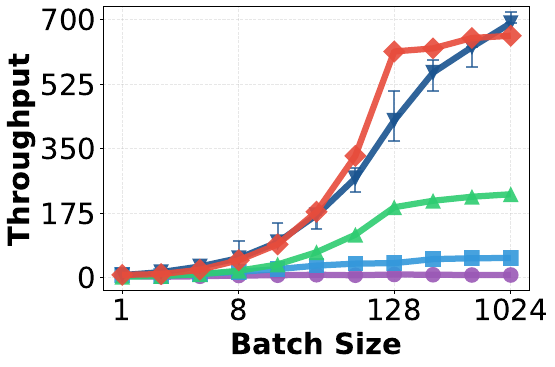}
  \end{subfigure}\hfill
  \begin{subfigure}[c]{0.24\textwidth}
    \includegraphics[width=\linewidth]
    {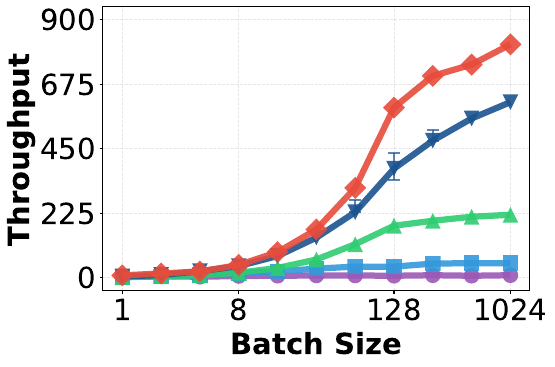}
  \end{subfigure}

  \begin{minipage}[c]{0.03\textwidth}
    \centering
    \rotatebox{90}{\small\textbf{Vocab=201088}}
  \end{minipage}%
  \hfill
  \begin{subfigure}[c]{0.24\textwidth}
    \includegraphics[width=\linewidth]
    {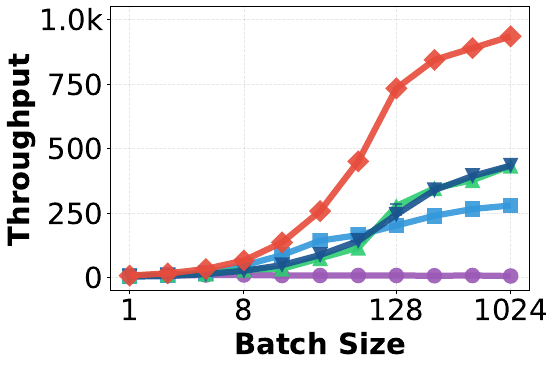}
  \end{subfigure}\hfill
  \begin{subfigure}[c]{0.24\textwidth}
    \includegraphics[width=\linewidth]
    {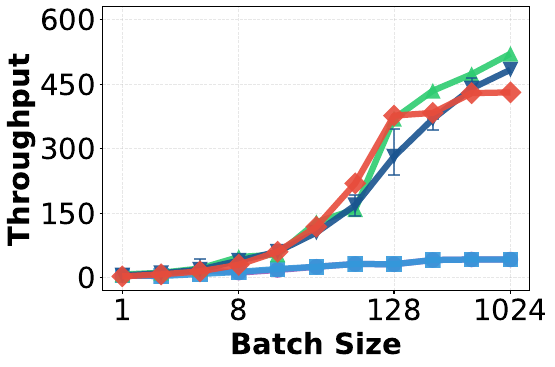}
  \end{subfigure}\hfill
  \begin{subfigure}[c]{0.24\textwidth}
    \includegraphics[width=\linewidth]
    {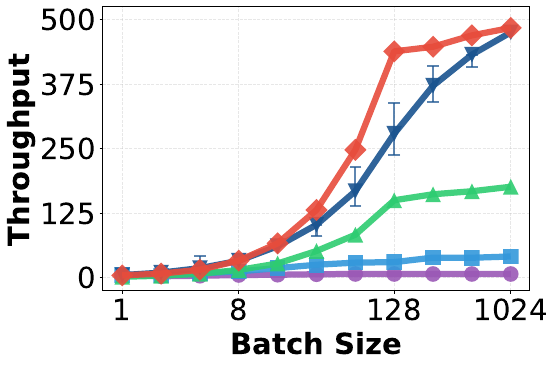}
  \end{subfigure}\hfill
  \begin{subfigure}[c]{0.24\textwidth}
    \includegraphics[width=\linewidth]
    {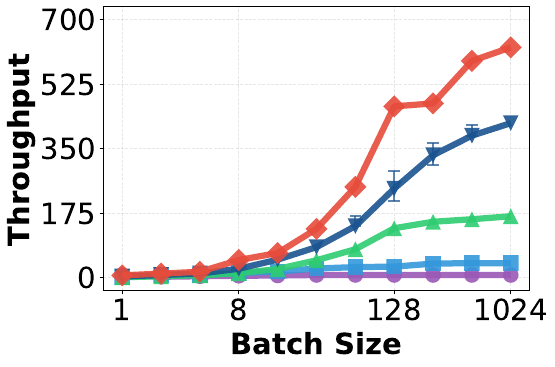}
  \end{subfigure}

  \begin{minipage}[c]{0.03\textwidth}
    \centering
    \rotatebox{90}{\small\textbf{Vocab=262208}}
  \end{minipage}%
  \hfill
  \begin{subfigure}[c]{0.24\textwidth}
    \includegraphics[width=\linewidth]
    {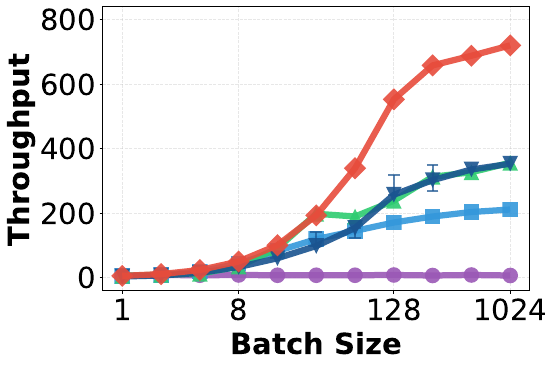}
  \end{subfigure}\hfill
  \begin{subfigure}[c]{0.24\textwidth}
    \includegraphics[width=\linewidth]
    {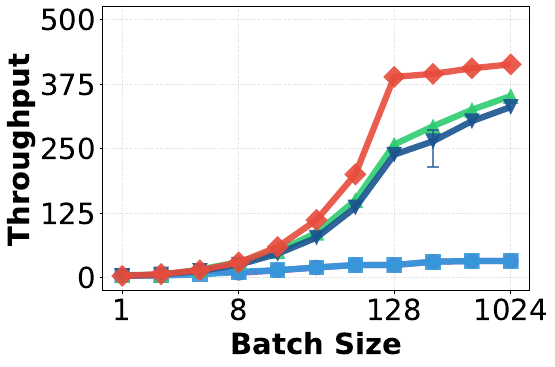}
  \end{subfigure}\hfill
  \begin{subfigure}[c]{0.24\textwidth}
    \includegraphics[width=\linewidth]
    {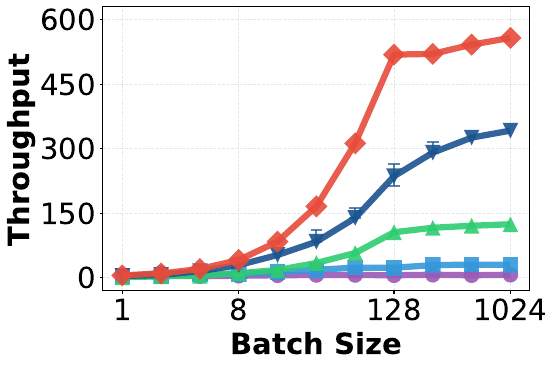}
  \end{subfigure}\hfill
  \begin{subfigure}[c]{0.24\textwidth}
    \includegraphics[width=\linewidth]
    {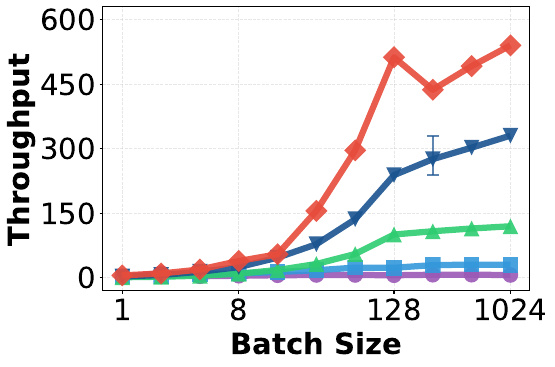}
  \end{subfigure}

  \includegraphics[width=0.65\textwidth]
  {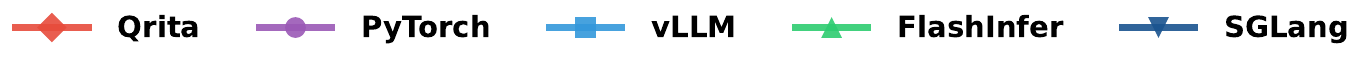}

  \caption{Sampling throughput (samples/ms) over various models and executions on an NVIDIA H100, averaged across 100 runs with a confidence interval of 96\%.}
  \label{fig:H100_Performance}
  \vspace{-5pt}
\end{figure*}

We present the execution throughput of \sys and the baselines on NVIDIA H100 in Figure~\ref{fig:H100_Performance}. We execute a batched logits tensor of shape $[\textit{Batch\_size}, \textit{Vocab\_size}]$, and divide the batch size by the execution time to get our throughput measurement. Overall, \sys outperforms all baselines in most configurations, with a few exceptions where SGLang or FlashInfer, which have relaxed output constraints with reduced functionality, achieve comparable throughput. As such, \sys can be considered as a direct improvement to the baselines in both performance and functionality.
Interestingly, the throughput of \sys scales linearly as batch size increases up to 128, and the throughput increase slows down after. This is because we allocate a Triton Program per multiprocessor of the GPU, of which the H100 has 132. As such, our current \sys implementation has a scalability limitation in the batch dimension. This can be solved by creating a \sys implementation where each Triton Program handles multiple batches in parallel, which we leave as future work.

\begin{figure*}[t]
    \centering

      \hspace*{0.02\textwidth}
  \begin{minipage}{0.24\textwidth}
    \centering\textbf{$k$=10, $p$=None}
  \end{minipage}\hfill
  \begin{minipage}{0.24\textwidth}
    \centering\textbf{$k$=None, $p$=0.7}
  \end{minipage}\hfill
  \begin{minipage}{0.24\textwidth}
    \centering\textbf{$k$=50, $p$=0.9}
  \end{minipage}\hfill
  \begin{minipage}{0.24\textwidth}
    \centering\textbf{$k$=Rand, $p$=Rand}
  \end{minipage}


    \begin{minipage}[c]{0.03\textwidth}
        \centering
        \rotatebox{90}{\small\textbf{RTX4090}}
    \end{minipage}%
    \hfill
    \begin{subfigure}[c]{0.24\textwidth}
        \includegraphics[width=\linewidth]
        {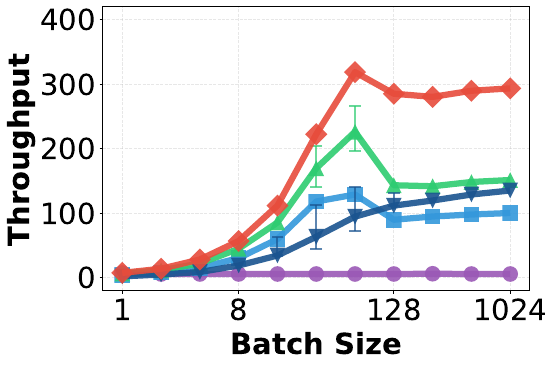}
    \end{subfigure}\hfill
    \begin{subfigure}[c]{0.24\textwidth}
        \includegraphics[width=\linewidth]
        {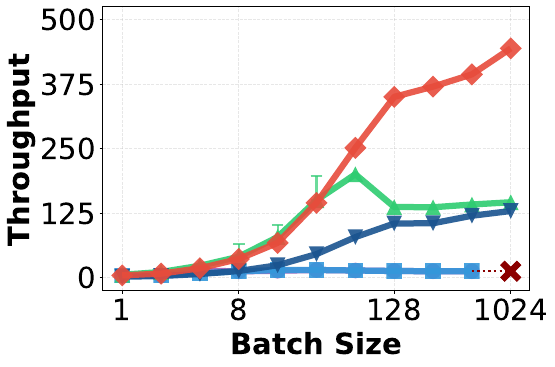}
    \end{subfigure}\hfill
    \begin{subfigure}[c]{0.24\textwidth}
        \includegraphics[width=\linewidth]
        {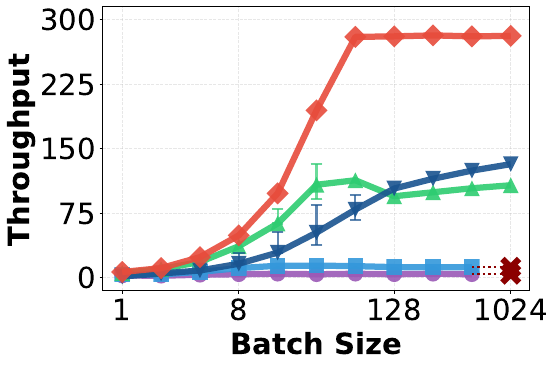}
    \end{subfigure}\hfill
    \begin{subfigure}[c]{0.24\textwidth}
        \includegraphics[width=\linewidth]
        {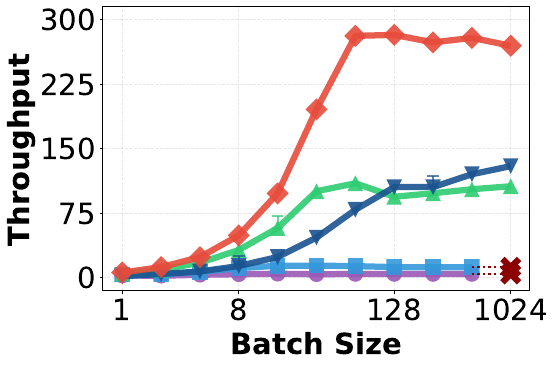}
    \end{subfigure}

    \begin{minipage}[c]{0.03\textwidth}
        \centering
        \rotatebox{90}{\small\textbf{MI350X}}
    \end{minipage}%
    \hfill
    \begin{subfigure}[c]{0.24\textwidth}
        \includegraphics[width=\linewidth]
        {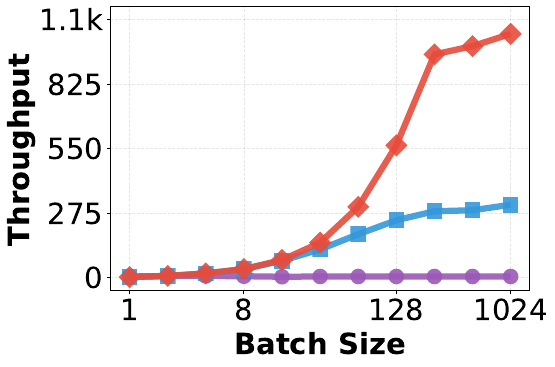}
    \end{subfigure}\hfill
    \begin{subfigure}[c]{0.24\textwidth}
        \includegraphics[width=\linewidth]
        {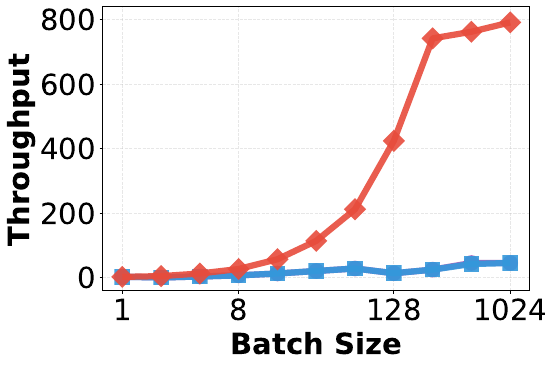}
    \end{subfigure}\hfill
    \begin{subfigure}[c]{0.24\textwidth}
        \includegraphics[width=\linewidth]
        {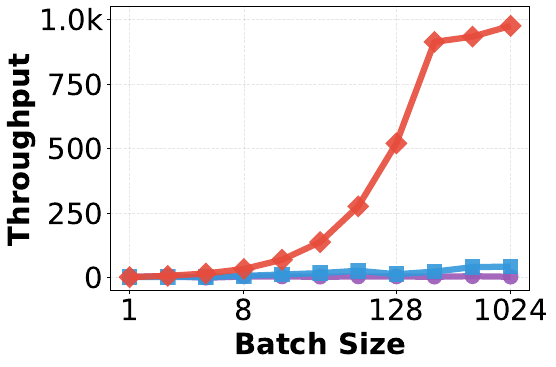}
    \end{subfigure}\hfill
    \begin{subfigure}[c]{0.24\textwidth}
        \includegraphics[width=\linewidth]
        {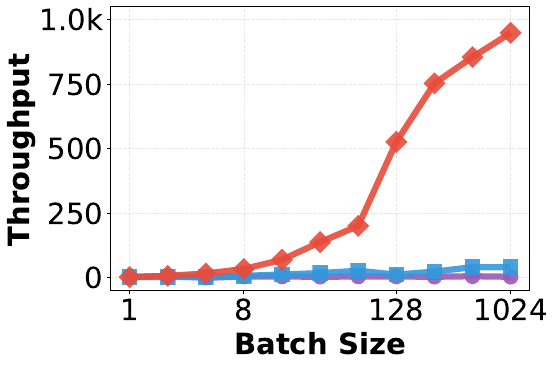}
    \end{subfigure}

    \includegraphics[width=0.85\textwidth]{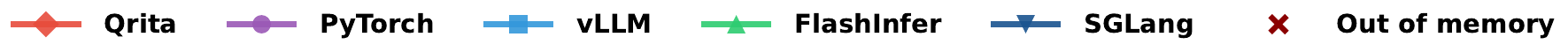}

    \caption{Sampling throughput (samples/ms) of Gemma3-4B over various executions on NVIDIA RTX4090 and AMD MI350X, averaged across 100 runs with a confidence interval of 96\%. High-performance samplers of FlashInfer and SGLang are written in CUDA and are unavailable on AMD.}
    \label{fig:4090_performance}
    \vspace{-16pt}
\end{figure*}

Figure~\ref{fig:4090_performance} shows the throughput results on  RTX4090 and AMD MI350X. For RTX4090, \sys achieves up to 2$\times$ the throughput of baselines even when compared to SGLang or FlashInfer. Considering that \topk and \topp are mostly memory-bound operations, we attribute this difference to the memory improvements of \sys through the use of search space truncation and the sorting-free search, performing favorably on RTX4090 with reduced memory performance. On MI350X, \sys is unmatched by all baselines, as \sys outclasses PyTorch and vLLM while the high-performance sampler implementations of FlashInfer and SGLang are available only on NVIDIA GPUs.

  
  


\subsection{Truncation Hit-rate and Memory Use}
\label{sec:mem_and_hitrate}

\begin{table}[htbp]
  \centering
  
  \begin{minipage}[t]{0.55\textwidth}
    \vspace{0pt} 
    \centering
    \small
    \begin{tabular}{lcccc} 
      \toprule
      & \multicolumn{2}{c}{\textbf{$k=50$}} & \multicolumn{2}{c}{\textbf{$p=0.9$}} \\
      
      \cmidrule(lr){2-3} \cmidrule(lr){4-5}
      
      \textbf{Model} & \textbf{$N_{t}$} & \textbf{Hitrate} & \textbf{$S_{t}$} & \textbf{Hitrate}  \\
      \midrule
      
      R1-Llama-8B & 276 & 1.0 & 0.984 & 0.985  \\
      Qwen3-8B & 399 & 1.0 & 0.967 & 0.901  \\
      GPT-OSS-20B & 286 & 1.0 & 0.952 & 0.938  \\
      Gemma3-4B & 2368 & 1.0 & 0.999 & 1.0  \\
      \bottomrule
    \end{tabular}
    \vspace{1pt}
    \caption{Gaussian $\sigma$-truncation hitrate with avg. number of outliers $N_t$ and avg. sum of outlier probabilities $S_t$.}
    \label{tab:hitrate}
  \end{minipage}\hfill
  \begin{minipage}[t]{0.43\textwidth}
    \vspace{0pt} 
    \centering
    \includegraphics[width=\linewidth]{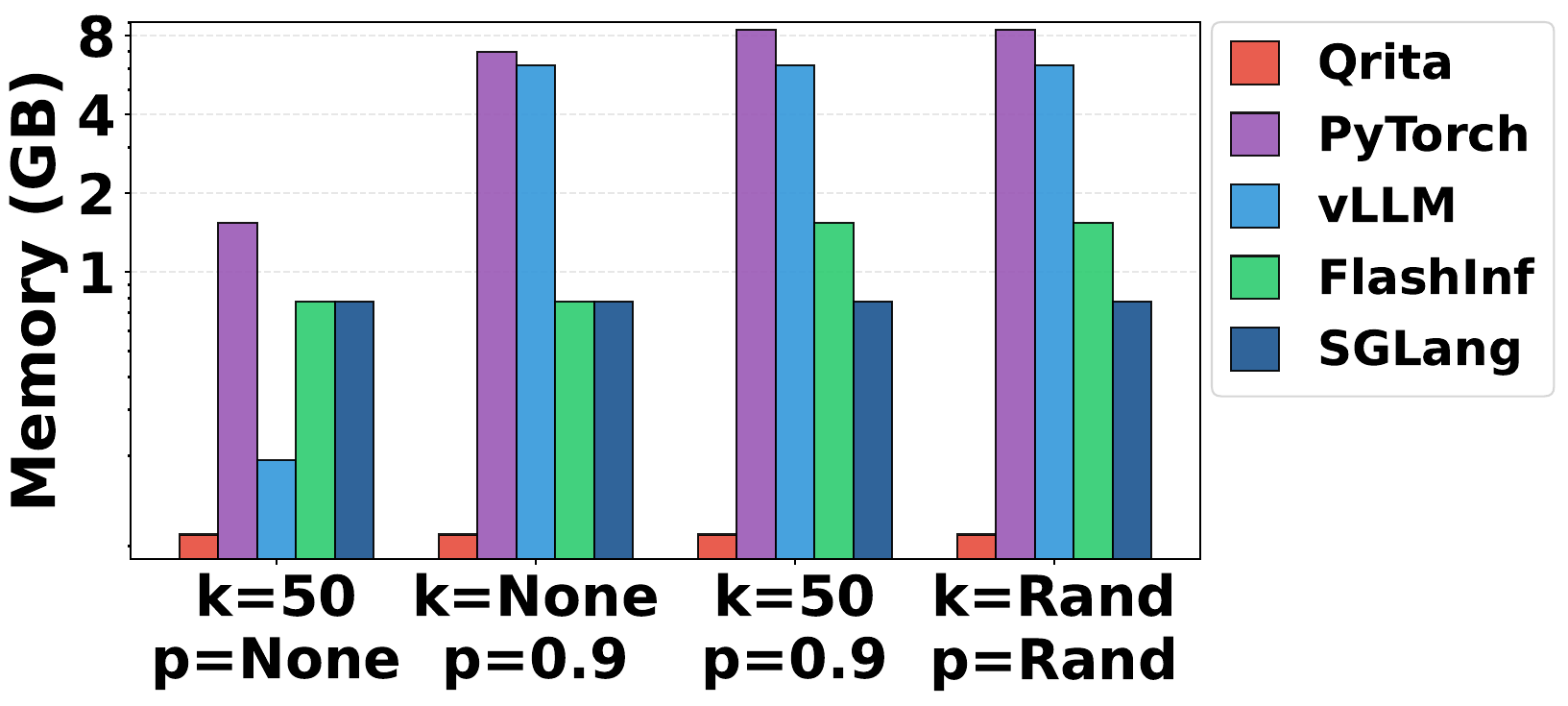}
    \captionof{figure}{Memory usage during sampling. Vocab size=201088, Batch size=1024}
    \label{fig:memory_use}
  \end{minipage}
  \vspace{-20pt}
\end{table}


Table~\ref{tab:hitrate} presents the hit-rate of the Gaussian $\sigma$-truncation, along with the average number of outliers and average sum of outlier probabilities of the truncated sets while running the Wikitext-2 dataset. We observe that the current truncation threshold for \topk is tuned to be extremely conservative, achieving a constant hit-rate of 1.0 and a headroom of hundreds of logits even though $k$ is set to only 50. 
It is possible to gain speedup by tightening the $\sigma$-truncation threshold, but we find the benefits marginal except for the Gemma3 case. This is because modern GPUs have enough thread resources to handle hundreds of logits in parallel, with minimal impact on latency. However, if we extend \sys to support multiple batches per Triton program as mentioned before, it is likely that a tighter threshold on each batch will be beneficial for efficient thread-level parallelism across the batch dimension.

For the \topp case, our results show that while our Gaussian-based modeling of the logits provides robust lower-bound truncation thresholds, larger variances in the sum of probability and hit-rate exist across models compared to the \topk case. This is because the inclusion of outlier values in the real logit distribution greatly alters the probability calculation of the softmax operator compared to pure Gaussian-based modeling. As such, to achieve a more accurate truncation, it may be beneficial to include a profiling step that creates a dedicated $\delta$ lookup table for each model.

Figure~\ref{fig:memory_use} presents the memory usage of \sys and baselines across different executions on a logarithmic scale. While \sys achieves lower memory usage compared to all baselines, it can also be implemented in-place to have \textit{zero memory use} by skipping Gaussian $\sigma$-truncation. However, we believe that the performance benefit from truncation is large enough to justify a buffer space, as shown in ablation studies. Alternatively, as the truncated set size is often a few thousand elements at most, the truncation buffer may be reduced from the current full vocabulary buffer to further optimize \sys memory use.

\subsection{End-to-end Performance}
\label{sec:e2e}

\begin{table*}[htbp]
  \centering
  \resizebox{\textwidth}{!}{
  \begin{tabular}{lccccccccc}
    \toprule
    \multirow{2}{*}{\centering Model} & 
    \multirow{2}{*}{\begin{tabular}[c]{@{}c@{}}Input/Output\\Tokens\end{tabular}} & 
    \multirow{2}{*}{\begin{tabular}[c]{@{}c@{}}Sampling\\Skipped\end{tabular}} &
    \multicolumn{2}{c}{$k$=50} & 
    \multicolumn{2}{c}{$p$=0.9} & 
    \multicolumn{2}{c}{$k$=50, $p$=0.9} & 
    \multirow{2}{*}{\begin{tabular}[c]{@{}c@{}}Average\\Speedup\end{tabular}} \\
    \cmidrule(lr){4-5} \cmidrule(lr){6-7} \cmidrule(lr){8-9}
    & & & \multicolumn{1}{c}{Baseline} & \multicolumn{1}{c}{\sys} & 
    \multicolumn{1}{c}{Baseline} & \multicolumn{1}{c}{\sys} & 
    \multicolumn{1}{c}{Baseline} & \multicolumn{1}{c}{\sys} & \\
    \midrule
    R1-Dist-Llama-8B & 256/1024 & 14.80 & 12.70 & 14.50 & 12.49 & 14.22 & 
    12.39 & 14.54 & \textbf{$1.15\times$} \\
    R1-Dist-Llama-8B & 1024/256 & 26.28 & 24.56 & 26.02 & 24.36 & 25.63 & 
    24.29 & 26.00 & \textbf{$1.06\times$} \\
    R1-Dist-Llama-8B & 1024/1024 & 14.21 & 12.78 & 14.03 & 12.61 & 13.76 & 
    12.56 & 14.03 & \textbf{$1.10\times$} \\
    Qwen3-8B & 256/1024 & 13.95 & 12.04 & 13.68 & 11.79 & 13.57 & 11.57 & 
    13.66 & \textbf{$1.16\times$} \\
    Qwen3-8B & 1024/256 & 25.75 & 23.22 & 25.50 & 23.08 & 25.23 & 22.95 & 
    25.40 & \textbf{$1.10\times$} \\
    Qwen3-8B & 1024/1024 & 13.29 & 11.83 & 13.09 & 11.67 & 12.98 & 11.62 & 
    13.11 & \textbf{$1.12\times$} \\
    GPT-OSS-20B & 256/1024 & 30.08 & 20.15 & 28.39 & 19.66 & 26.43 & 19.32 & 
    28.55 & \textbf{$1.41\times$} \\
    GPT-OSS-20B & 1024/256 & 52.13 & 43.85 & 51.72 & 43.31 & 47.72 & 42.63 & 
    52.10 & \textbf{$1.17\times$} \\
    GPT-OSS-20B & 1024/1024 & 35.16 & 25.84 & 33.50 & 25.21 & 31.32 & 24.73 & 
    33.67 & \textbf{$1.30\times$} \\
    Gemma3-4B & 256/1024 & 15.63 & 12.01 & 15.41 & 11.67 & 15.27 & 11.51 & 
    15.53 & \textbf{$1.31\times$} \\
    Gemma3-4B & 1024/256 & 35.45 & 29.28 & 34.09 & 28.72 & 33.85 & 28.49 & 
    34.11 & \textbf{$1.18\times$} \\
    Gemma3-4B & 1024/1024 & 19.29 & 14.83 & 18.28 & 14.48 & 18.21 & 14.30 & 
    18.42 & \textbf{$1.26\times$} \\
    \bottomrule
  \end{tabular}
  }
  \caption{End-to-end serving throughput of baseline vLLM, vLLM with sampling kernel skipped (theoretical max performance), and vLLM with \sys on NVIDIA H100 (KTokens/sec).}
  \label{tab:sampling_performance}
\end{table*}

\begin{table*}[htbp]
  \centering
  \resizebox{\textwidth}{!}{
  \begin{tabular}{lcclccccccccc}
    \toprule
    \multirow{2}{*}{Model} & \multirow{2}{*}{$k$} & \multirow{2}{*}{$p$} & \multirow{2}{*}{Impl} &
    \multicolumn{3}{c}{TTFT (ms)} & \multicolumn{3}{c}{TPOT (ms)} & \multicolumn{3}{c}{E2EL (ms)} \\
    \cmidrule(lr){5-7} \cmidrule(lr){8-10} \cmidrule(lr){11-13}
    & & & & p50 & p95 & p99 & p50 & p95 & p99 & p50 & p95 & p99 \\
    \midrule
    Qwen3.5-35B-A3B & 50 & 0.9 & Baseline & 77.24 & 114.55 & 145.80 & 21.90 & 23.51 & 24.20 & 5512.51 & 6064.16 & 6090.64 \\
    Qwen3.5-35B-A3B & 50 & 0.9 & \sys & 71.86 & 123.28 & 132.75 & 19.28 & 20.51 & 20.87 & 4882.88 & 5288.63 & 5322.34 \\
    \midrule
    Qwen3.5-35B-A3B & 50 & none & Baseline & 78.10 & 114.45 & 134.24 & 22.05 & 23.56 & 23.97 & 5491.71 & 6068.22 & 6086.06 \\
    Qwen3.5-35B-A3B & 50 & none & \sys & 77.15 & 136.00 & 163.68 & 20.56 & 22.25 & 24.63 & 5196.33 & 5727.92 & 5739.14 \\
    \midrule
    Qwen3.5-35B-A3B & none & 0.9 & Baseline & 75.96 & 114.73 & 140.62 & 22.09 & 23.64 & 23.73 & 5631.94 & 6106.15 & 6125.16 \\
    Qwen3.5-35B-A3B & none & 0.9 & \sys & 75.95 & 147.27 & 179.28 & 20.63 & 22.11 & 22.27 & 5254.81 & 5705.91 & 5738.00 \\
    \bottomrule
  \end{tabular}
  }
  \caption{End-to-end serving latency of baseline vLLM and vLLM with \sys on NVIDIA H100.}
  \label{tab:serving_latency}
\end{table*}

Table~\ref{tab:sampling_performance} reports the end-to-end serving performance improvement of \sys by comparing the serving throughput of baseline vLLM execution and vLLM with \sys as its \topk and \topp sampler. We also report the theoretical max throughput when the \topk and \topp sampling kernel is entirely skipped. As can be seen in the table, \sys{} consistently outperforms the vLLM baseline, delivering up to 1.41$\times$ speedup and often \textbf{providing performance within 5\% of the theoretical max values}. 

We note that the speedup is affected by the ratio between active model parameters and vocabulary size. For example, GPT-OSS-20B only has 3B active parameters, being an MoE model, and therefore achieves the best speedup among the tested models. Similarly, Qwen3-Next-80B-A3B~\cite{yang2025qwen3technicalreport} with 3B active parameters may be expected to have roughly 2.7$\times$ the performance improvement compared to Qwen3-8B, or up to 40\% end-to-end throughput improvement similar to GPT-OSS-20B. Therefore, while the end-to-end benefits of \sys are largely model-dependent, it can \textit{significantly improve the inference performance} of small to mid-sized dense LLMs, or parameter-efficient MoE models.

Comparing the \sys and vLLM's time-to-first-token (TTFT), time-per-output-token (TPOT), and end-to-end latency (E2EL) results in Table~\ref{tab:serving_latency}, we see that there are cases where \sys delivers a slower p95/p99 tail latency for TTFT and TPOT. However, \sys consistently delivers lower average (p50) latencies, resulting in around an 5-13\% reduction in end-to-end latencies even in p95/p99 cases.

\subsection{Ablation Studies}
\label{sec:ablation}

\begin{table*}[t]
    \centering
    \fontsize{8}{9}\selectfont
    \setlength{\tabcolsep}{4.5pt}
    \begin{tabular}{c | c c c | c c c | c c | c}
        \toprule
         & Trunc. Hit & Trunc. Fallback & No Trunc. & Quaternary & Binary & Sort  & Dup. & Autotune & Time (ms) \\
        \hline
        Run A & O & X & X & O & X & X & O & O & 0.919 \\
        Run B & O & X & X & O & X & X & O & X & 1.354 \\
        \hline
        Run C & O & X & X & O & X & X & X & X & 1.308 \\
        Run D & X & O & X & O & X & X & X & X & 5.373 \\
        Run E & X & X & O & O & X & X & X & X & 4.931 \\
        \hline
        Run F & O & X & X & X & O & X & X & X & 1.281 \\
        Run G & X & O & X & X & O & X & X & X & 7.773 \\
        Run H & X & X & O & X & O & X & X & X & 7.274 \\
        \hline
        Run I & O & X & X & X & X & O & X & X & 2.099 \\
        Run J & X & X & O & X & X & O & X & X & 16.487 \\
        \bottomrule
    \end{tabular}
    \caption{Ablation study of \sys techniques. $k$=50, $p$=0.9, Vocab size=128256, Batch size=1024}
    \label{tab:ablation_chart}
    \vspace{-14pt}
\end{table*}

In Table~\ref{tab:ablation_chart}, we provide ablation studies on the various techniques of \sys. Run A is the full \sys execution, which includes all key techniques introduced in Section~\ref{sec:method}. 
We first study the effect of the \textbf{Tiered Triton Block Size Autotune} by comparing Run A with Run B, which uses only a single block size of 4096 for all vectorized operators. Autotuning reduces the execution latency by almost 30\%, proving to be a critical optimization for \sys that operates on vectors of varying sizes.

The \textbf{overhead of duplication handling} is shown by comparing Run B with Run C, which excludes the duplicate logit removal logic from the output mask generation. Latency is slightly improved by about 3.5\%. We include duplication handling for the sake of correctness, but if performance is critical and a more relaxed output constraint can be tolerated, duplication handling can be removed.

Runs C, D, and E compare \textbf{the effect of Gaussian $\sigma$-truncation} by comparing the time taken when the truncation hits, when it misses, and when it is excluded. The truncation delivers the biggest speedup of all techniques used, providing 74\% latency reduction compared to the excluded case that must search the full logits set. By comparing Runs D and E, we also observe that the overhead of the truncation is about 0.4ms, which is quite significant when compared to fully optimized Run A, contributing 48\% of the total runtime. As such, reducing the overhead of Gaussian $\sigma$-truncation may be a good future direction for \sys.

By comparing Runs F, G, and H against Runs C, D, and E, we observe \textbf{the impact of quaternary search over binary search} for pivot finding. Surprisingly, quaternary search is about 2\% slower than binary search for the truncation hit case, as the reduced search space minimizes the impact of search iterations. However, due to the large absolute latency reduction of 2.3ms it provides in the fallback case, we keep quaternary search as our default approach. 

Comparing Runs I and J against Runs C and E, we also observe \textbf{the benefit of quaternary search over the sorting-based algorithms}, where it provides about 38\% and 70\% reduction in latency, proving the benefit of the more memory-efficient search-based algorithm over the sorting-based algorithm. Note that the configuration of Run J is identical to the naive PyTorch baseline.

\subsection{Empirical Exactness}
\label{sec:exactness}

We use the \topk and \topp sampler test suite of vLLM\footnote{\url{https://github.com/vllm-project/vllm/blob/main/tests/v1/sample/test_topk_topp_sampler.py}} to compare \sys against the PyTorch sort-based reference. The tests sweep across different batch sizes, vocabulary sizes, and $k$ and $p$ values, including requests where \topk or \topp is skipped for part of the batch. They also cover edge cases: extreme $k$ ($k=1$ and $k=\mathcal{V}$), extreme $p$ ($p$ near $0$ and $p=1$), identical logits, near-identical logits (spread of $10^{-6}$), and $-\infty$ logits (e.g., from structured output masks), including checks that no NaN is produced. \sys passes all tests, giving a match rate of 100\% and a miss rate of 0\% against the PyTorch reference on this suite.

%% file: 5.Discussions.tex
\section{Discussions}


\textbf{Comparison to Radix \topk, Radix \topp, and other GPU \topk and \topp Algorithms: }
\sys is composed of three distinct components serving different purposes. The component comparable to Radix \topk and \topp or other GPU \topk and \topp algorithms is the Quaternary Pivot Search, as it is the main ``exact \topk and \topp'' algorithm. On the other hand, the goal of Gaussian $\sigma$-truncation is \textit{to reduce the \topk and \topp search space in a single pass}, and the goal of Duplication Handling is to \textit{ensure deterministic results} in the presence of identical values, which is not directly comparable to other GPU \topk and \topp algorithms, as they add on top of an exact \topk and \topp algorithm.


We chose the search-based approach for the ``exact \topk and \topp'' component, as it is the easiest to fold the detection and counting of duplicate logits into the algorithm. Radix \topk and \topp, RadiK, and many high-performance GPU \topk and \topp algorithms are inefficient or non-deterministic in the presence of duplicate values compared to our Quaternary Search, which can detect the pivot falling on duplicate values. Our choice of Quaternary Search is based on the effectiveness it brings to duplicate handling, and not because it is necessarily better than radix-based algorithms in isolation.

\textbf{Computational Complexity of \sys: } 
The computational complexity of the Gaussian $\sigma$-truncation is $\mathcal{O}(N)$ as it passes twice over the logit set of size $N$. If the truncated set size is $M$, the complexity of quaternary search is $\mathcal{O}(M\log(M))$, as we need $\log(M)$ passes on the logit set of size $M$. As $M \ll N$ ($k$ or $p$ $\ll vocab\_size$), the complexity of Qrita is $\mathcal{O}(N)$.
However, RadixSelect or Quickselect may outperform quaternary search with large $M$, as they also achieve $\mathcal{O}(M)$ by reducing the search space using gather on each step.
We do not apply such optimization because (1) we limit $M$ using $\sigma$-truncation, and (2) gathering operations on GPUs are very inefficient. Current quaternary search performs scan operations using parallel threads, which are mapped to contiguous reads. However, gather operations require random read and write operations, which limit batching of memory accesses and add the overhead of write accesses, causing higher latency per step. This is why we limit the gather, or the search space reduction, to only the Gaussian $\sigma$-truncation pass. 

\textbf{Exactness Analysis: }
Despite the empirical correctness shown using a test suite in Section~\ref{sec:exactness}, it is possible that \sys may mismatch some PyTorch sort-based \topk and \topp references in theory.

\textit{(a) Gaussian $\sigma$-truncation.} We gather outlier logits strictly larger than the threshold $t$. Therefore, the outlier set of size $M$ must contain the top-$M$ values of the logit set. If $M \geq k$, the outlier set is a superset of the \topk set. The same holds for \topp if the sum of the outlier probabilities is larger than $p$. If the outlier set does not satisfy this property, we fall back to the full logit set. Thus, the truncated set is guaranteed to be a superset of the \topk or \topp set, and \sys is correct if the remaining search finds the exact set within it.

\textit{(b) Quaternary pivot search.} The search is correct if its resolution, or its ability to differentiate between two adjacent values, is finer than the gap between the two adjacent values of the pivot.

\textit{(b-1) Absolute resolution.} The absolute resolution is the tolerance of $10^{-9}$ (roughly $2^{-30}$). Logits near the cutoff pivot that differ by less than $10^{-9}$ may be treated as identical (duplicate) values. However, this is unlikely in practice, as logits are typically FP16 or BF16 values upcast to FP32. The finest resolution of FP16 is $2^{-24} \approx 6.0 \times 10^{-8}$, which is coarser than $10^{-9}$. For BF16, with its 7-bit mantissa, a resolution finer than $10^{-9}$ only occurs for $|z| < 2^{-22} \approx 2.4 \times 10^{-7}$. With reasonable $k$ and $p$ values and real model predictions, pivots rarely land in such near-zero values.

\textit{(b-2) Relative resolution.} Since we reduce the search range by $4\times$ per iteration for up to 18 iterations, the relative resolution is $4^{-18} \approx 1.5 \times 10^{-11}$ times the original search range. Logits near the pivot that differ by less than this may lead to an incorrect pivot. For probabilities, which lie within $[0, 1]$, the relative resolution is at most $1.5 \times 10^{-11}$. For logits, it is finer than the absolute resolution of $10^{-9}$ whenever the search range is below about 68, which is the case for typical logit distributions.

\textit{(c) Duplication handling.} We keep the first-occurrence set, which is \textit{equivalent to a stable sort-based \topk and \topp} that preserves the order of identical elements. However, \textbf{PyTorch's sort is unstable by default}, with undefined order between identical elements, unless the \texttt{stable} flag is set. Therefore, \sys results may differ from PyTorch's default sort. However, this is not necessarily incorrect, and we believe that equivalence to a stable sort is the most reasonable design choice for determinism.

%% file: 6.Conclusion.tex
\section{Conclusion}
\label{sec:conclusion}

This paper proposes \sys, an efficient \topk and \topp truncation algorithm for GPUs. Our method supports both \topk and \topp in $\mathcal{O}(n)$ while preserving the precise output semantics of both algorithms. We implement \sys in Triton and show that \sys achieves up to $1.4\times$ higher throughput against state-of-the-art baselines, providing a high-performance solution for LLM decoding.  

\newpage

\begin{ack}
    We would like to thank Woosuk Kwon, Yilong Zhao, Mayank Mishra, the vLLM team, and the UC Berkeley Sky Computing Lab for sharing their valuable insights in this research.
    
    This research was supported by the Amazon AI PhD Fellowship awarded to Jongseok Park, as well as NSF awards IIS-1955488, IIS-2027575, DOE awards DE-SC0016260, AC02-05CH11231, DARPA Agreement No. HR00112590131, and gifts to the UC Berkeley Sky Computing Lab from Accenture, Algorithmic SuperIntelligence Labs, Amazon, AMD, Anyscale, Broadcom, cmpnd, Google, IBM, Intel, Intesa Sanpaolo, Lambda, Lightspeed, Mirendil, NVIDIA, Soma Labs, and VESSL.
\end{ack}

